\documentclass[letterpaper, journal]{IEEEtran}
\usepackage{amsmath,amsfonts}
\usepackage{array}
\usepackage{textcomp}
\usepackage{stfloats}
\usepackage{url}
\usepackage{verbatim}
\usepackage{graphicx}
\usepackage{cite}
\usepackage{xcolor}
\usepackage{subcaption}
\usepackage{mathtools}
\usepackage{amssymb}
\usepackage{tabulary}
\usepackage{booktabs}
\usepackage[ruled,linesnumbered]{algorithm2e}
\usepackage{hyperref}
\usepackage{setspace}
\usepackage{bm}
\usepackage{multirow}
\usepackage{makecell}
\usepackage{adjustbox}

\newcommand{\sss}{\scriptscriptstyle}

\newcommand{\B}[1]{\boldsymbol{\mathbf{#1}}}

\newcommand{\defeq}{\vcentcolon=}

\newcommand{\skewsym}[1]{[#1]_{\times}}

\newcommand{\nbx}[1]{{}^N{#1}_{B}}
\newcommand{\nnx}[1]{{}^N{#1}_{N}}
\newcommand{\bbx}[1]{{}^B{#1}_{B}}
\newcommand{\half}{\frac{1}{2}}

\newcommand{\nonds}{\B{f}_{\sss\B{\mathcal{N}}}}
\newcommand{\worldds}{\B{f}_{\sss\B{\mathcal{W}}}}

\title{
Global-State-Free Obstacle Avoidance for Quadrotor Control in Air-Ground Cooperation
}

\author{
	\IEEEauthorblockN{
        Baozhe Zhang\IEEEauthorrefmark{2}\textsuperscript{2}, 
        Xinwei Chen\IEEEauthorrefmark{2}\textsuperscript{1, 2}, 
		Qingcheng Chen\textsuperscript{3},
        Chao Xu\textsuperscript{1, 2},
        Fei Gao\textsuperscript{1, 2},
        and Yanjun Cao\textsuperscript{1, 2}
        }
        \vspace{-1.0cm}
}

\begin{document}

\maketitle

\begingroup
\renewcommand\thefootnote{\IEEEauthorrefmark{2} }
\footnotetext{\textbf{Equal contribution}}
\renewcommand\thefootnote{}
\footnotetext{This work was supported by National Nature Science Foundation of China under Grant 62103368. (Corresponding author: Yanjun Cao, Chao Xu. \tt \{yanjunhi, cxu\}@zju.edu.cn)}
\renewcommand\thefootnote{\textsuperscript{1} }
\footnotetext{State Key Laboratory of Industrial Control Technology, Institute of Cyber-Systems and Control, Zhejiang University, Hangzhou, 310027, China.}
\renewcommand\thefootnote{\textsuperscript{2} }
\footnotetext{Huzhou Institute of Zhejiang University, Huzhou, 313000, China.}
\renewcommand\thefootnote{\textsuperscript{3} }
\footnotetext{Shanghai Institute of Special Equipment Inspection \& Technical Research, Shanghai, 200062, China.}

\endgroup

\begin{abstract}
	CoNi-MPC~\cite{zhang2023coni} provides an efficient framework for UAV control in air-ground cooperative tasks by relying exclusively on relative states, eliminating the need for global state estimation. However, its lack of environmental information poses significant challenges for obstacle avoidance. To address this issue, we propose a novel obstacle avoidance algorithm, \textbf{Co}operative \textbf{N}on-\textbf{i}nertial frame-based \textbf{O}bstacle \textbf{A}voidance (CoNi-OA), designed explicitly for UAV-UGV cooperative scenarios without reliance on global state estimation or obstacle prediction.
	CoNi-OA uniquely utilizes a single frame of raw LiDAR data from the UAV to generate a modulation matrix, which directly adjusts the quadrotor's velocity to achieve obstacle avoidance. This modulation-based method enables real-time generation of collision-free trajectories within the UGV's non-inertial frame, significantly reducing computational demands (less than 5 ms per iteration) while maintaining safety in dynamic and unpredictable environments.
	The key contributions of this work include: (1) a modulation-based obstacle avoidance algorithm specifically tailored for UAV-UGV cooperation in non-inertial frames without global states; (2) rapid, real-time trajectory generation based solely on single-frame LiDAR data, removing the need for obstacle modeling or prediction; and (3) adaptability to both static and dynamic environments, thus extending applicability to featureless or unknown scenarios.

\end{abstract}

\begin{IEEEkeywords}
	Obstacle Avoidance, Air-Ground Cooperation, Non-Inertial Model, Modulated Dynamical System
\end{IEEEkeywords}

\section{Introduction}

\IEEEPARstart{M}{ulti-robot} cooperation systems, such as UAV-UGV pairs \cite{miki2019UAVUGV}, leader-follower systems \cite{di2021leaderfollower}, multi-agent formations \cite{quan2022formation}, and autonomous landing platforms \cite{falanga2017landing} have been extensively utilized in surveillance, search-and-rescue operations, and cinematography.
In CoNi-MPC \cite{zhang2023coni}, the authors proposed an air-ground robot system in which the UAV (or quadrotor) is actively controlled to fulfill a task along with an independently controlled UGV.
It eliminates the need for SLAM and continuous trajectory re-planning, offering a computationally efficient and robust approach that adapts seamlessly to dynamic and unpredictable environments, making it broadly suitable for diverse air-ground cooperative applications.
However, CoNi-MPC does not consider the influence of obstacles in the environment, and the UAV cannot guarantee its own safety during flight, strongly limiting its application.
In this letter, we propose a novel obstacle avoidance method, \textbf{Co}operative \textbf{N}on-\textbf{i}nertial frame based \textbf{O}bstacle \textbf{A}voidance (CoNi-OA) for quadrotor control, to address the safety issue in UAV-UGV collaboration system.

\begin{figure}[!t]
	\setlength{\belowcaptionskip}{-0.25cm}
	\centering
	\includegraphics[width=.476\textwidth]{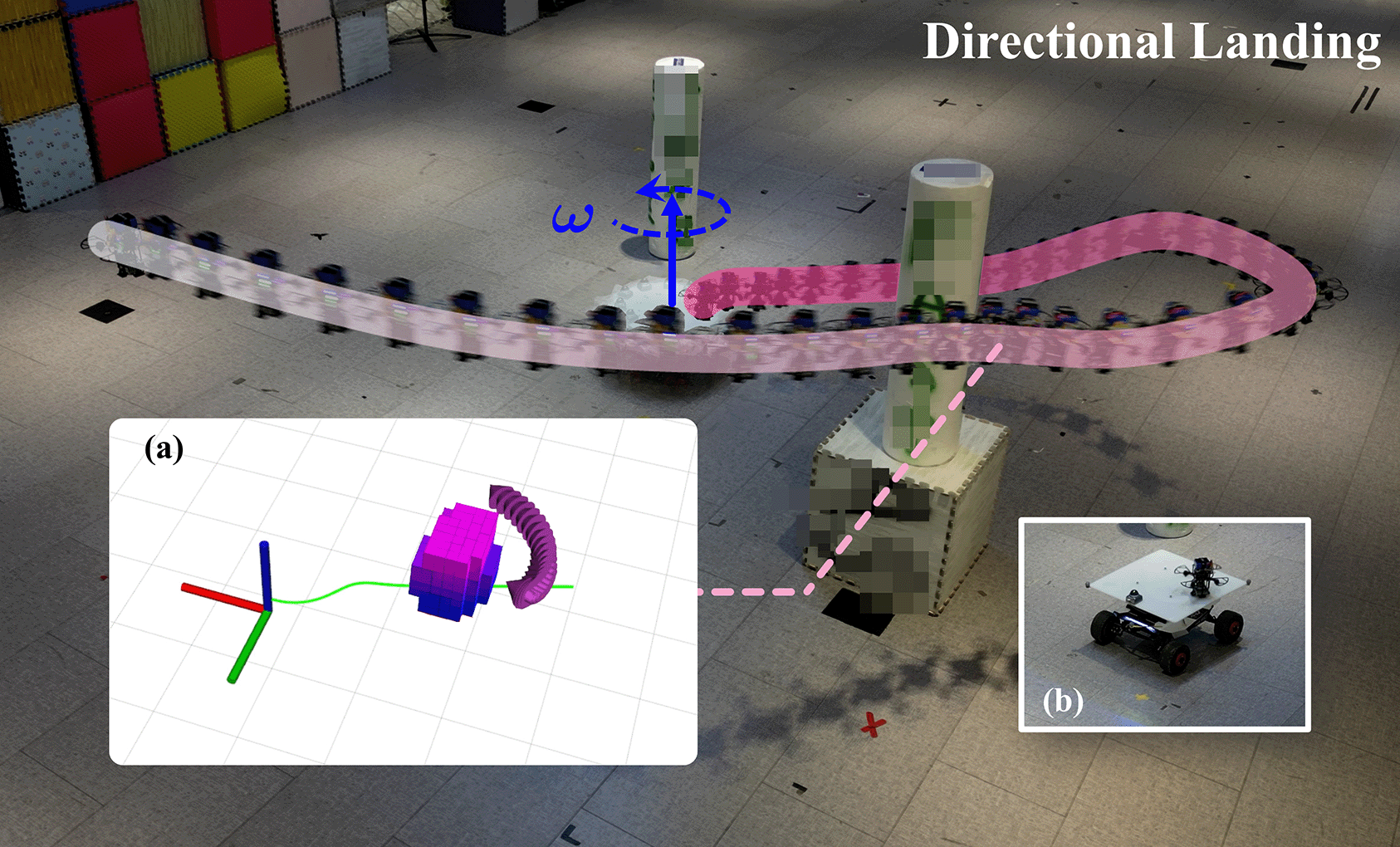}
	\caption{A quadrotor avoids obstacles and lands on a rotating UGV from a fixed direction by applying CoNi-OA, in which the pre-generated collision-free trajectory (green) is obstructed during landing. (a) shows the experiment from a third-person view in the world frame where the quadrotor avoides the ``dynamic'' obstacle in the UGV's non-inertial frame;
		(b) shows the final successful landing.}
	\label{fig:landing}
\end{figure}

Directly controlling the UAV in the UGV's non-inertial frame provides a more intuitive and efficient way to perform air-ground cooperative tasks by removing the dependence on the global state of the UAV in the world frame.
Within this cooperative framework, directly controlling the UAV enables it to execute cooperative tasks such as leader-follower control, directional landing, and orbit flight, without the need to both predict UGV motion and continuously replan global trajectories.
Considering that the global state estimation in the world frame depends on either extra infrastructure (GPS, motion capture systems, etc.) or featured environments (visual or geometry features, etc.), control in the UGV's non-inertial frame greatly expands the application scenario to feature-less or unknown environments.
However, it also introduces new challenges for obstacle avoidance function which is unavoidable in real-world applications.
For example, the UAV must avoid obstacles in the UGV's non-inertial reference frame, and the static obstacles in the inertial world reference frame become ``\emph{dynamic}''\footnote{
	In this work, we define ``dynamic'' obstacles as obstacles whose positions change unpredictably in the UGV's non-inertial frame due to the UGV's unknown motion. Conversely, obstacles that remain fixed relative to the UGV are considered ``static''.
} and unpredictable in the non-inertial frame due to the unknown UGV's movements.
Therefore the UAV must be able to avoid both static and dynamic obstacles in real time while maintaining its dynamic feasibility in the non-inertial frame.
There are three main challenges to address in this context:
\begin{itemize}
	\item \textbf{Avoiding unpredictable dynamic obstacles}: Obstacles in the non-inertial frame become dynamic and unpredictable due to the unknown UGV's motion. The UAV need to effectively avoid these dynamic obstacles while following the planned relative motion to the UGV.
	\item \textbf{Rapid response capability}: Achieving rapid obstacle avoidance is crucial, particularly for environments with dynamic obstacles. Methods based on real-time optimization often rely on static assumptions like Euclidean Signed Distance Fields (ESDFs) or flying corridors, where the update of them can lag in dynamic environments. Delays in updating these ESDFs or constraints hinder their responsiveness and may lead to collisions.
	\item \textbf{Quadrotor dynamic feasibility}: Ensuring that the generated trajectories comply with the UAV's dynamic constraints is essential for safe and smooth flight. This is particularly challenging in the non-inertial frame for air-ground cooperation, where the UAV must account for the influence of unknown UGV motion on its own dynamics.
\end{itemize}
To overcome these challenges, we propose CoNi-OA method as a local planner for the quadrotor with a LiDAR sensor mounted on it to avoid dynamic obstacles in the UGV's non-inertial frame with only one frame of LiDAR sampling in each control iteration.
Inspired by \cite{huber2023fast}, we employ a sample-based modulation matrix (directly from LiDAR sampled data points) to modulate the initial velocity from a high-level planner, iteratively generating local collision-free trajectories for the quadrotor that adhere to the non-inertial dynamical system (DS) of the quadrotor.
Different from \cite{huber2023fast} that applying the modulated DS to a point-mass model in 2D space, CoNi-OA is designed for a rigid body like a quadrotor in 3D space and in the non-inertial frame, which supports more aggressive maneuvers than the point-mass model.
The generated trajectories are then effectively tracked by CoNi-MPC, enabling the quadrotor to cooperatively avoid obstacles in dynamic scenarios.

\subsection*{Novelty and Contribution}
The contributions of this paper are summarized as follows:
\begin{itemize}
	\item \textbf{First non-inertial frame obstacle avoidance for UAV-UGV cooperation}: Unlike existing methods requiring global states or planning in the world frame\cite{zhou2020ego,paolo1998vo,van2011avo}, we propose the first modulation-based solution specifically designed for UAV control in a UGV's non-inertial frame. This eliminates dependency on global positioning systems or SLAM while handling both static and dynamic obstacles through a LiDAR sensor.

	\item \textbf{Single-frame reactive modulation}: Compared to optimization-based approaches\cite{zhou2020ego} that require obstacle modeling/prediction, our method directly processes raw LiDAR points to compute velocity modulation matrices in $<5$ ms per iteration. This enables collision avoidance without prior knowledge of obstacle motion patterns.

	\item \textbf{General Framework for Obstacle Avoidance}: The proposed method can serve as a general framework for obstacle avoidance in both static and dynamic environments, which not only facilitates avoidance capability for quadrotors within the air-ground system but is also adaptable for degenerated scenarios that control quadrotors in the world frame.

\end{itemize}

\section{Related Work}

Obstacle avoidance is one of the main function for motion planning in robot autonomous navigation.
Well-studied high-level motion planners focus on finding collision-free paths from the robot's current position to the global goal, mostly in static environments.
Classical techniques, such as rapidly-exploring random tree (RRT), probabilistic roadmap (PRM), and their variants \cite{lavalle2001rapidly, kavraki1996probabilistic, karaman2011sampling}, use sampling in the configuration space to generate paths and rely on replanning to adapt to dynamic changes.
As these methods assume static obstacles, making it difficult to navigate in highly dynamic environments.

State-of-the-art real-time optimization-based methods have gained significant attention by applying trajectory optimization or optimized control following the planned path from the high-level planner.
Model Predictive Control (MPC) can generate collision-free trajectories by incorporating obstacle-related terms into the cost function or constraints \cite{park2009obstacle, lindqvist2020nmpc, li2023robocentric}. However, these methods require a trade-off between time horizon length, computational load, and optimality, with shorter horizons often leading to locally optimized solutions. For trajectory optimization approaches, ESDFs \cite{ratliff2009chomp, kalakrishnan2011stomp, ding2019efficient, zhou2021raptor} and safe corridors \cite{deits2015efficient, gao2020teach, wang2022geometrically} are commonly used to model the environment and generate collision-free trajectories.
Ego-Planner \cite{zhou2020ego} introduces an optimization-based method that avoids using ESDFs, retrieving obstacle information only when necessary.
However, these methods generally assume static obstacles during the optimization process, which may lead to collisions in dynamic environments. Additionally, when applied to dynamic settings, frequent updates of the obstacle information such as ESDFs or safe corridors are required, introducing delays that can lead to collisions.

Reactive obstacle avoidance strategies, such as velocity obstacles (VO) \cite{paolo1998vo, van2008rvo, van2011avo}, are designed for dynamic environments. VO computes collision-free movements in velocity space by estimating the velocities and accelerations of moving obstacles, considering their current motion and potential future positions. \cite{van2008rvo, van2011avo} extend VO to reciprocal velocity obstacles (RVO) and acceleration-velocity obstacles (AVO), incorporating higher-order motion information. However, these methods rely on accurate estimates of obstacle velocities and accelerations, which necessitate precise identification from raw sensor data. Without this information, VO may struggle to respond to dynamic changes, particularly in noisy or incomplete sensor environments. While effective in dynamic settings, VO can still be prone to local minima due to its lack of long-term planning, especially in static environments.

Furthermore, the modulated dynamical system inspired by harmonic potential fields offers a more coupled method for collision-free navigation in complex environments \cite{huber2019avo, huber2022avo, huber2023fast}. \cite{huber2019avo, huber2022avo} modulate an initial linear system to achieve convergence around convex obstacles and to reach a global goal. Nevertheless, these methods require prior or online shape analysis of obstacles, which can introduce delays in dynamic environments. On the other hand, \cite{huber2023fast} uses a sample-based approach to modulate the linear system with LiDAR data, avoiding time-consuming analysis. But it is limited to the point-mass model in 2D space and doesn't consider a rigid body like a quadrotor, or to a quadrotor in the non-inertial frame.

\section{Preliminaries}

In \cite{zhang2023coni}, the authors developed a cooperative system consisting of a UGV and a UAV, capable of executing various interactive tasks including leader-following, directional landing, orbit flight, etc. To ensure the safety of the UAV flight while carrying out these complex interactions, we consider a modulation-based obstacle avoidance method.

\subsection{Quadrotor Dynamics in Non-Inertial Frame}

The state of the quadrotor system in the non-inertial frame is defined as $\B{x} = [\nbx{\B{p}};\nbx{\B{v}};\nbx{\B{q}}] \in \mathbb{R}^{6} \times \mathbb{S}^3$ where $\nbx{\B{p}}$, $\nbx{\B{v}}$, and $\nbx{\B{q}}$ respectively denote relative position, velocity, and orientation (unit quaternion) between the UAV's body frame $B$ and the UGV's body frame $N$.
The input of the system is defined as $\B{u} = [T;\bbx{\B{\Omega}}]\in\mathbb{R}^4$, where $T=\sum_{i=1}^{4}T_i$ represents the mass-normalized collective thrust of the four propellers and $\bbx{\B{\Omega}}$ is the angular velocity (body rate) of the quadrotor's body frame.
The relative system flow in \cite{zhang2023coni} can be summarized as:
\begin{equation}
	\begin{aligned}
		\label{eq:relative_dynamics}
		\nbx{\dot{\B{p}}} & = \nbx{\B{v}}                                                                                                               \\
		\nbx{\dot{\B{v}}} & = -\skewsym{\nnx{\B{\beta}}} \nbx{\B{p}} - 2\skewsym{\nnx{\B{\Omega}}} \nbx{\dot{\B{p}}}                                    \\
		                  & -\skewsym{\nnx{\B{\Omega}}}^2 \nbx{\B{p}} + \nbx{\B{q}} \odot \bbx{\B{T}} \odot \nbx{\B{q}}^{-1} - \nnx{\B{a}}^\texttt{IMU} \\
		\nbx{\dot{\B{q}}} & = -\half\nnx{\B{\Omega}} \odot \nbx{\B{q}} + \half \nbx{\B{q}} \odot \bbx{\B{\Omega}}
	\end{aligned}
\end{equation}
where $\nnx{\B{a}}^\texttt{IMU}$ is the UGV's IMU acceleration measurement, $\nnx{\B{\Omega}}$ is the body rate of the UGV, $\nnx{\B{\beta}}$ is the angular acceleration of the UGV in its body frame, $\left[\cdot\right]_{\times}:\mathbb{R}^3 \rightarrow \mathbb{R}^{3\times3}$ is the skew-symmetric matrix operator, and $\odot$ is the Hamilton product operator.
We denote the dynamical system (DS) in Eq.~\refeq{eq:relative_dynamics} as $\nonds: \mathbb{R}^{6}\times\mathbb{S}^3\times\mathbb{R}^4\rightarrow\mathbb{R}^{10}$, where $\B{\mathcal{N}} = [\nnx{\B{a}}^\texttt{IMU};\nnx{\B{\Omega}};\nnx{\B{\beta}}]\in\mathbb{R}^9$ are the non-inertial quantities that represent a time-varying parameter set of the system dynamics.
Conceptually, if $\B{\mathcal{W}} = \B{\mathcal{N}} = [(0,0,g);\B{0};\B{0}]$ where $g = 9.8 \;\text{m}/\text{s}^2$ is the gravity, then the system dynamics $\nonds$ will degenerate to the quadrotor dynamics in the world inertial frame $\worldds$.

\subsection{Obstacle Avoidance Through Modulated Dynamical System}

\begin{figure}[!t]
	\setlength{\belowcaptionskip}{-0.25cm}
	\centering
	\includegraphics[width=.476\textwidth]{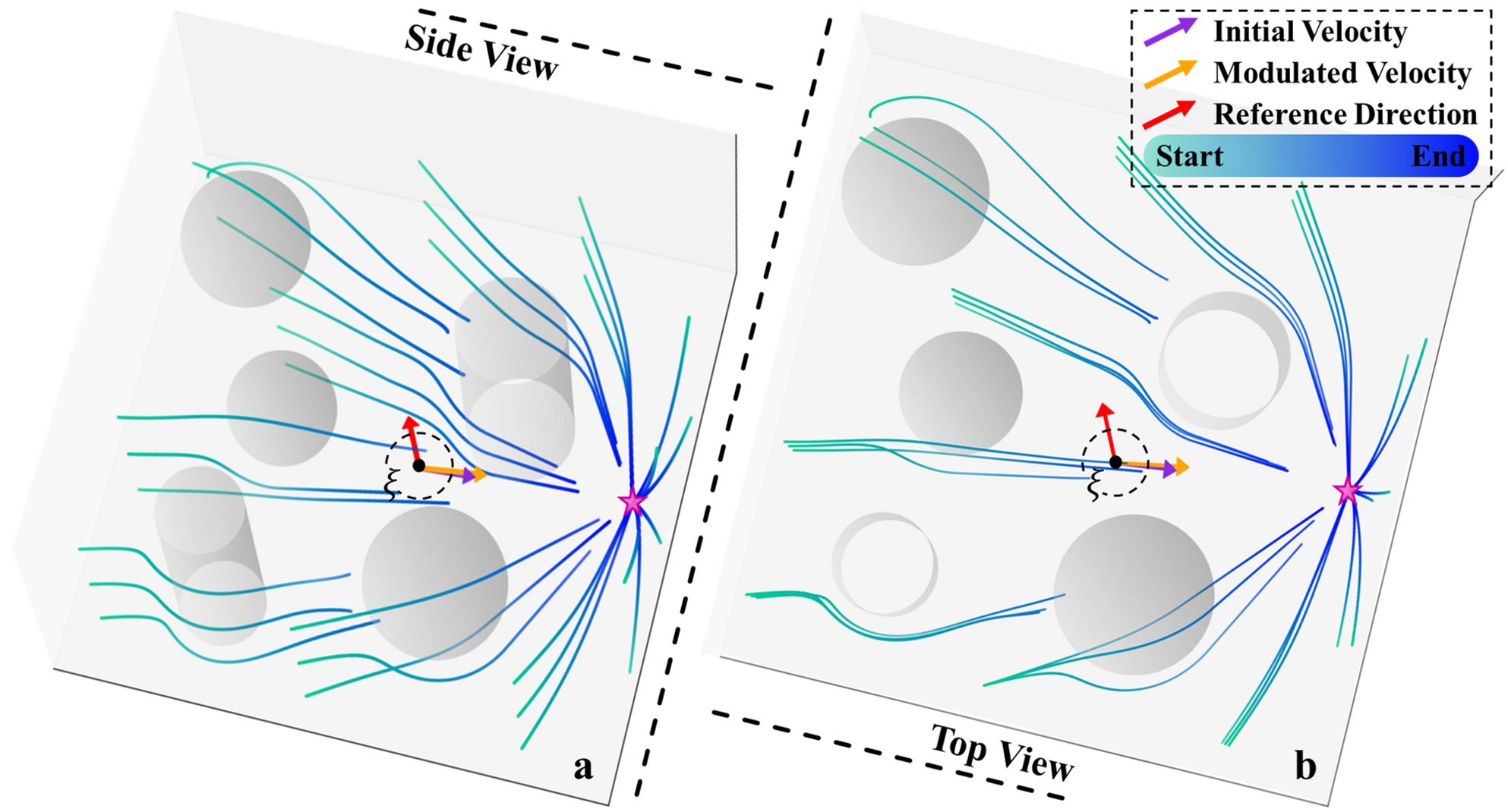}
	\caption{The point-mass model's collision-free navigation around obstacles using the modulated DS. Red arrow indicate the summed reference direction from obstacles at position $\B{\xi}$, with the purple and orange arrows showing the direction of the initial and modulated velocities, respectively.}
	\label{fig:flow}
\end{figure}

For obstacle avoidance, the study \cite{huber2023fast} presents a general approach, Fast Obstacle Avoidance (FOA), to generate a modulated DS, which is characterized by the application of a modulation matrix $\B{\mathcal{M}} \in \mathbb{R}^{d \times d}$ to the point-mass DS:
\begin{equation}
	\label{eq:general_modulated_ds}
	\dot{\B{\xi}} = \B{\mathcal{M}}(\B{\xi}, \B{v})\B{v}, \;
	\B{\mathcal{M}}(\B{\xi}, \B{v}) = \B{E}(\B{\xi})\B{D}(\B{\xi}, \B{v})\B{E}(\B{\xi})^{-1}
\end{equation}
where $\B{\xi} \in \mathbb{R}^d$ is the robot's position, $\B{v} \in \mathbb{R}^d$ is the velocity from a high-level planner, $\B{E}(\B{\xi})$ is the basis matrix, and $\B{D}(\B{\xi})$ is the eigenvalue matrix that defines the scaling along each direction. When $\B{\mathcal{M}}$ has full rank, the modulated DS will maintain its asymptotic stability without introducing new extrema \cite{kronander2015incremental}. Fig.~\ref{fig:flow} shows the collision-free motion of the point-mass modulated DS as described in Eq.~\ref{eq:general_modulated_ds}.

A way of applying the modulated DS to obstacle avoidance for quadrotor flight is to incorporate the point-mass model to generate trajectories:
\begin{equation}
	\label{eq:mass_point_modulation}
	\B{v}_{\text{mod}} \defeq \B{\mathcal{M}} \B{v}_{\text{init}}
\end{equation}
where the point-mass model serves as an intermediary between a higher-level trajectory planner that generates $\B{v}_{\text{init}}$ and a standard MPC controller that operates based on the model $\nonds$, which is fed with $\B{v}_{\text{mod}}$.
However, this may raise the question regarding the suitability of the point-mass model for a rigid body such as a quadrotor, or even for a quadrotor in a non-inertial frame.
To enable obstacle avoidance for the quadrotor flight in the non-inertial frame while maintaining its dynamic feasibility, we propose a method to iteratively generate local collision-free reference trajectories using the modulated DS and employ CoNi-MPC to track these trajectories.

\section{Sample-Based Modulation for Cooperative Obstacle Avoidance}

\begin{figure}[!t]
	\setlength{\belowcaptionskip}{-0.25cm}
	\centering
	\includegraphics[width=.476\textwidth]{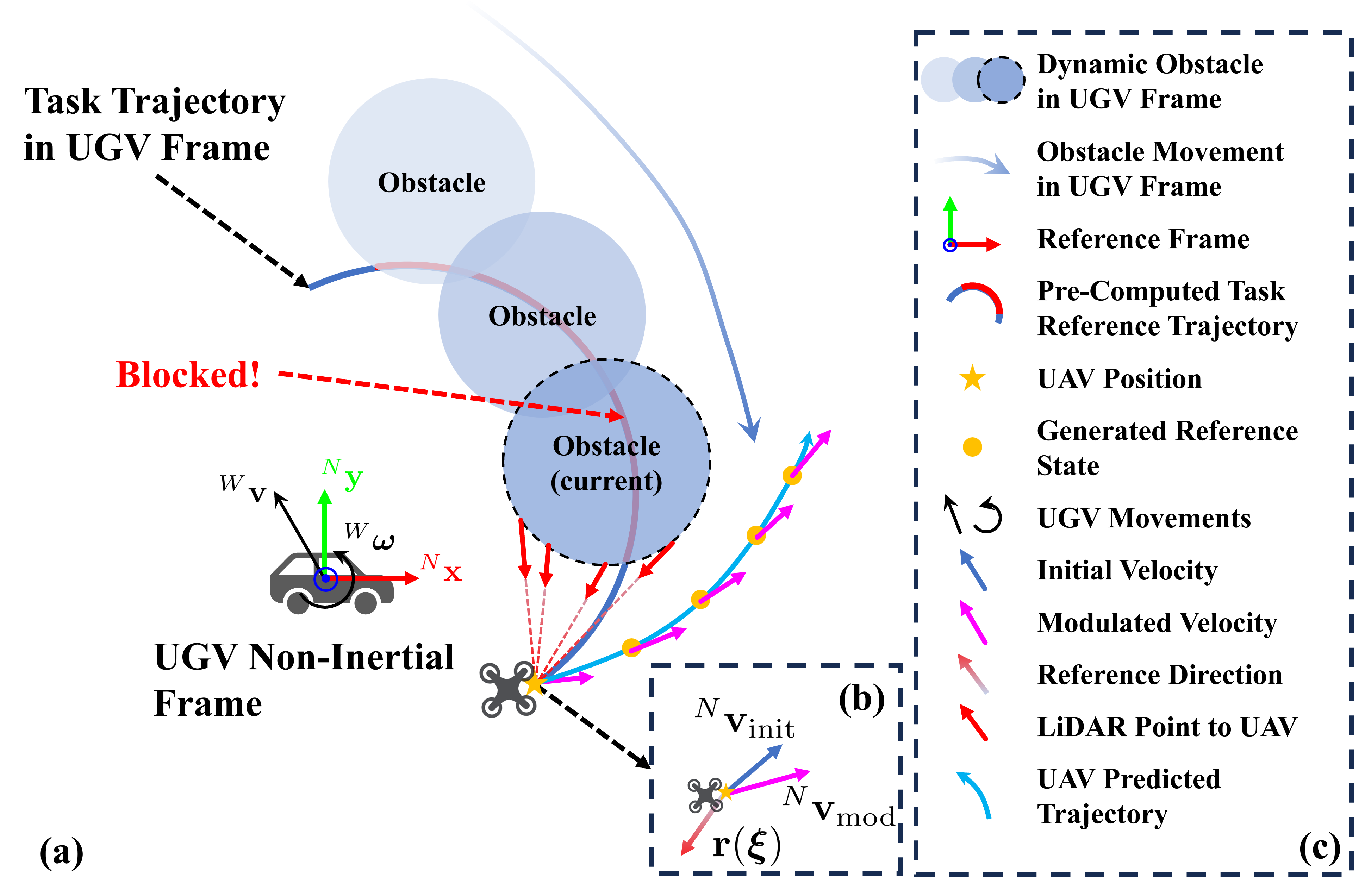}
	\caption{
		CoNi-OA adds obstacle avoidance to CoNi-MPC\cite{zhang2023coni} for robust UAV-UGV collaboration.
		(a) The UAV, operating in the UGV's non-inertial reference frame with pre-computed task trajectory which can be blocked by dynamic obstacles, uses lidar data to compute safe trajectories while adapting to the UGV's unpredictable motion.
		(b) By modulating the initial velocity on the pre-computed task trajectory, the UAV generates collision-free trajectories toward the original task trajectory.
		(c) shows the labels.
	}
	\label{fig:concept}
\end{figure}

CoNi-OA adopts the sample-based technique to generate the modulation matrix $\B{\mathcal{M}}$ (\ref{subsec: Sampled-Based Modulation Matrix}) described in \cite{huber2023fast}. It should be noted that the sampled data points are collected from the LiDAR sensor mounted on the UAV and they need to be represented in the non-inertial frame $N$. With the modulation matrix $\B{\mathcal{M}}$, the modulated DS outlined in Eq.~\ref{eq:mass_point_modulation} can be used to generate collision-free trajectories (\ref{subsec: Modulation-Based Collision-Free Trajectory Generation}) while obeying the rigid-body DS in Eq.~\ref{eq:relative_dynamics}. CoNi-MPC will track these trajectories, enabling the quadrotor to cooperatively avoid obstacles during specific UAV-UGV interaction tasks (\ref{subsec: Implementation}).

\subsection{Sampled-Based Modulation Matrix \label{subsec: Sampled-Based Modulation Matrix}}

Since the UGV's frame $N$ is a non-inertial moving frame, most obstacles represented in this frame are dynamic and unpredictable.
As a result, we do not assume any prior knowledge about the obstacles' descriptions, but rather incorporate a sample-based method to perceive the obstacle as a set of data points in frame $N$ at runtime. This set is denoted by $\B{\xi}_o \in \mathbb{R}^3,o=1,\dots,n$, where $n \in \mathbb{R}_{>0}$ is the total number of points.
Each data point $\B{\xi}_o$ can be viewed as a spherical obstacle with a small radius $\delta$, so the obstacle space can be represented as $\B{\Phi}(t) = \bigcup_{o=1}^n\{ \B{\xi} : \lVert \B{\xi} - \B{\xi}_o \rVert \leq \delta \} $.

\begin{algorithm}[t]
	\caption{GenBasisMatrix}
	\label{algo:genBasis}
	\SetKwInput{Input}{Input}
	\SetKwInput{Output}{Output}
	\SetAlgoLined
	\Input{Weighted reference direction $\B{r}(\B{\xi})$}
	\Output{Basis matrix $\B{E}(\B{\xi})$}

	$\B{e}_1 \leftarrow  [1,0,0]^\top$

	$\B{r}(\B{\xi})_n \leftarrow {\B{r}(\B{\xi})}/{\lVert \B{r}(\B{\xi}) \rVert}$

	\tcp{Ensure $\B{w}$ and $\B{r}(\B{\xi})_n$ non-collinear}
	\eIf {$\B{r}(\B{\xi})_n^x > 0$}
	{$\B{w} \leftarrow \B{r}(\B{\xi})_n + \B{e}_1$}
	{$\B{w} \leftarrow \B{r}(\B{\xi})_n - \B{e}_1$}

	$\B{E}(\B{\xi}) \leftarrow \B{I} - {2\B{w}\B{w}^\top}/{\B{w}^\top\B{w}}$

	\tcp{Ensure $\B{r}(\B{\xi})$ and $\B{E}(\B{\xi})$ aligned}
	\If{$\B{r}(\B{\xi}) \B{E}(\B{\xi}) \B{e}_1 < 0$}
	{$\B{E}(\B{\xi}) \leftarrow -\B{E}(\B{\xi})$}
\end{algorithm}

Inspired by FOA, we formulate the modulation matrix
$\B{\mathcal{M}}$ in a manner similar to FOA but extended to the 3D space.
The first step involves computing the weighted reference direction $\B{r}(\B{\xi})$ based on the sampled points $\B{\Phi}(t)$ and the quadrotor's current relative position $\B{\xi} = \nbx{\B{p}}$:
\begin{equation}
	\label{eq:weighted_reference_direction}
	\B{r}(\B{\xi}) = \sum_{\B{\xi}_o \in \B{\Phi}}w_o(\B{\xi})\frac{\B{\xi}-\B{\xi}_o}{\lVert \B{\xi}-\B{\xi}_o \rVert}
\end{equation}
The weights $w_o(\B{\xi})$ are assigned as follows:
\begin{equation}
	\label{eq:weight}
	w_o(\B{\xi}) =
	\begin{cases}
		\hat{w}_o(\B{\xi}) / \hat{w}^{\text{sum}} & \text{if } \hat{w}^{\text{sum}} > 1 \\
		\hat{w}_o(\B{\xi})                        & \text{otherwise}
	\end{cases}
\end{equation}
with $\hat{w}^{\text{sum}} = \min(\sum_{o=1}^n \hat{w}_o(\B{\xi}), w^{\text{max}})$, and $w^{\text{max}} \in \mathbb{R}_{>0}$ is the max weight. $\hat{w}_o(\B{\xi})$ for each sampled point is evaluated as:
\begin{equation}
	\label{eq:distance}
	\hat{w}_o(\B{\xi}) = \left(  \frac{D^{scal}}{D_o(\B{\xi})} \right)^s ,
	D_o(\B{\xi}) = \lVert \B{\xi}-\B{\xi}_o \rVert - R
\end{equation}
where $D_o(\B{\xi})$ is the distance from a data point to the quadrotor, $R \in \mathbb{R}_{>0}$ is the radius of a sphere enclosing the quadrotor, $D^{\text{scal}} \in \mathbb{R}_{>0}$ and $s \in \mathbb{R}_{>0}$ represent the distance scaling parameters.
By calculating the weighted reference direction, we assume a \textit{virtual weighted} obstacle with its center placing along $\B{r}(\B{\xi})$.

The weighted reference direction $\B{r}(\B{\xi})$ can be viewed as the normal direction of the virtual obstacle pointing towards the quadrotor.
To generate the basis matrix $\B{E}(\B{\xi})$ with one of its axis parallel with $\B{r}(\B{\xi})$, we use Householder Transformation on $\B{I} - 2{\B{v}\B{v}^T}/{\|\B{v}\|^2}$, as described in Algo.~\ref{algo:genBasis}.
Note that $\B{E}(\B{\xi})$ is a symmetric and orthonormal matrix, i.e., $\B{E}(\B{\xi})^\top=\B{E}(\B{\xi})^{-1}=\B{E}(\B{\xi})$.
The initial velocity from a high-level planner can be mapped to the space where $\B{E}(\B{\xi})$ represents, and modulated on normal and tangent directions by $\B{D}(\B{\xi})$ in Eq.~\ref{eq:general_modulated_ds}.

\begin{figure}[!t]
	\setlength{\belowcaptionskip}{-0.25cm}
	\centering
	\includegraphics[width=0.476\textwidth]{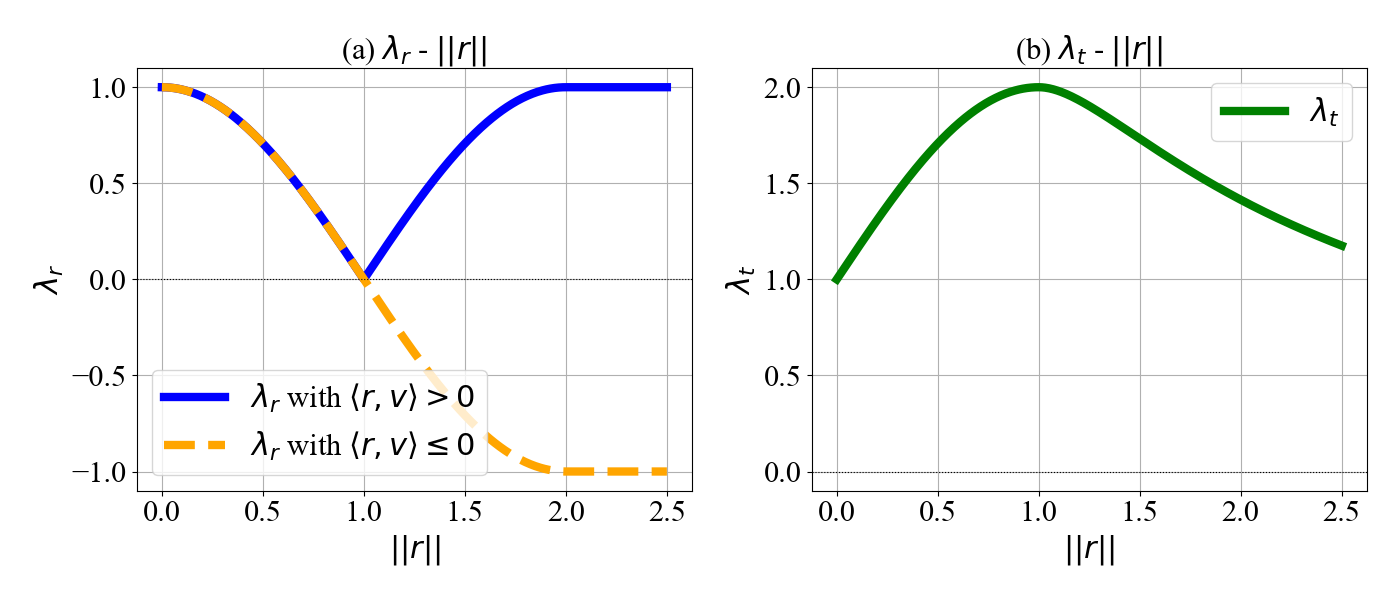}
	\caption{Visualization of $\lambda_r$ and $\lambda_t$ with respect to $\lVert \B{r}(\xi)\rVert$.}
	\label{fig:lambda_vs_r}
\end{figure}

The diagonal matrix $\B{D}(\B{\xi}) = diag(\lambda_r(\B{\xi}), \lambda_t(\B{\xi}), \lambda_t(\B{\xi}))$ contains the eigenvalues $\lambda_r(\B{\xi})$ and $\lambda_t(\B{\xi})$ of the modulation matrix $\B{\mathcal{M}}$, which respectively scales the initial velocity along the normal and tangential directions.
The visualization of $\lambda_r(\B{\xi})$ and $\lambda_t(\B{\xi})$ is illustrated in Fig.~\ref{fig:lambda_vs_r}.
To ensure safe obstacle avoidance, the scaling factor $\lambda_r(\B{\xi})$ of the normal direction needs to change the velocity's normal components (away from the virtual obstacle) when $\lVert \B{r}(\B{\xi}) \rVert$ is large, and maintain it when $\lVert \B{r}(\B{\xi}) \rVert$ is small.
Meanwhile, $\lambda_r(\B{\xi})$ needs to be smooth enough to avoid velocity glitching, which is defined as:
\begin{equation}
	\label{eq:lambda_r}
	\lambda_r(\B{\xi}) =
	\begin{cases}
		\cos(\frac{\pi}{2}\lVert\B{r}(\B{\xi})\rVert) & \text{if } \lVert \B{r}(\B{\xi}) \rVert < 2 \\
		-1                                            & \text{otherwise }
	\end{cases}
\end{equation}
If the angle between the initial velocity $\B{v}$ and the reference direction $\B{r}(\B{\xi})$ is acute and the quadrotor is close to the virtual obstacle, the sign of $\lambda_r(\B{\xi})$ needs to be positive to maintain the velocity direction:
\begin{equation}
	\label{eq:lambda_r_change_sign}
	\lambda_r(\B{\xi}) \leftarrow -\lambda_r(\B{\xi})\;\text{if } \langle \B{r}(\B{\xi}), \B{v} \rangle > 0\;\text{and } \lVert\B{r}(\B{\xi})\rVert > 1
\end{equation}
Similarly, the eigenvalues in the tangential direction are given by
\begin{equation}
	\label{eq:lambea_t}
	\lambda_t(\B{\xi}) =
	\begin{cases}

		1 + \sin(\frac{\pi}{2}\lVert\B{r}(\B{\xi})\rVert) & \text{if } \lVert \B{r}(\B{\xi}) \rVert < 1 \\
		2\sin(\frac{\pi}{2\lVert\B{r}(\B{\xi})\rVert})    & \text{otherwise }
	\end{cases}
\end{equation}
This ensures as the quadrotor approaches the virtual obstacle, its tangential velocity component increases, helping it bypass the obstacle. If the quadrotor gets too close, $\lambda_t(\B{\xi})$ approaches 0, causing it to be more inclined to move away from the obstacle along the normal direction rather than bypass it.

As the quadrotor moves away from an obstacle, its initial velocity, still modulated by
$\B{\mathcal{M}}$, is not desirable since the unconstrained velocity would already be adequate for collision avoidance. To improve this situation, $\lambda_r(\B{\xi})$ and $\lambda_t(\B{\xi})$ are designed as follows:
\begin{equation}
	\label{eq:lambda_r_t}
	\begin{aligned}
		\widetilde{\lambda}_t(\B{\xi}) & = ps + (1 - ps)\lambda_t(\B{\xi})                                                       \\
		\widetilde{\lambda}_r(\B{\xi}) & = \text{sgn}(s)p\widetilde{\lambda}_t(\B{\xi}) + (1 - \text{sgn}(s)p)\lambda_r(\B{\xi})
	\end{aligned}
\end{equation}
The weights are given as:
\begin{equation}
	\label{eq:weights}
	p = \min{\left(1, \frac{1}{\lVert\B{r}(\B{\xi})\rVert}\right)},
	s = \max{\left(0, \frac{\langle\B{r}(\B{\xi}), \B{v}\rangle}{\lVert\B{r}(\B{\xi})\rVert \lVert\B{v}\rVert}\right)^c}
\end{equation}
where $c \in \mathbb{R}_{>0}$ is the power parameter ($c=2$ in this paper).

\subsection{Modulation-Based Collision-Free Trajectory Generation \label{subsec: Modulation-Based Collision-Free Trajectory Generation}}

Different from the point-mass velocity modulation model in Eq.~\ref{eq:mass_point_modulation} from \cite{huber2023fast} in 2D, we design CoNi-OA to align with the quadrotor dynamics presented in Eq.~\ref{eq:relative_dynamics} to iteratively generate local collision-free trajectories for CoNi-MPC.

Given a global goal $\B{p}_{\text{goal}} \in \mathbb{R}^3$, we can employ proportional control based on a reference position $\B{p}_{\text{ref}}$ to generate the initial velocity $\B{v}_{\text{init}}$ for the quadrotor:
\begin{equation}
	\label{eq:v_init}
	\B{v}_\text{init} = bound(-k_p(\B{p}_{\text{ref}} - \B{p}_{\text{goal}}))
\end{equation}
where
$
	bound(\B{x}) = \lVert \B{x} \rVert \ge 1.0 \;?\; \frac{\B{x}}{\lVert \B{x} \rVert} : \B{x}
$
which forces the norm of the input vector $\lVert \B{x} \rVert \le 1.0$.
At this point $\B{p}_{\text{ref}}$, the reference velocity is determined by modulating the initial velocity $\B{v}_\text{init}$, denoted as $\B{v}_{\text{ref}} = \B{\mathcal{M}}(\B{p}_{\text{ref}}, \B{v}_\text{init})\B{v}_\text{init}$.
To derive the reference orientation at this point, we first calculate the modulated acceleration, which is obtained from the difference between the current and the last modulated velocities:
\begin{equation}
	\label{eq:acc_ref}
	\B{a}_{\text{ref}} =
	\frac{\B{v}_{\text{ref}} - \B{v}_{\text{last}} }{ \Delta t}
\end{equation}

Assuming the non-inertial quantities $\B{\mathcal{N}}$ as established in Eq.~\ref{eq:relative_dynamics} constant for each planning iteration and taking into account the differential flatness property attributed to quadrotors as proposed in \cite{mellinger2011minimum}, we define
\begin{equation}
	\begin{aligned}
		\label{eq:modulated_acc}
		\B{t} & = \B{a}_\text{ref} - (-\skewsym{\nnx{\B{\beta}}} \B{p}_\text{ref} - 2\skewsym{\nnx{\B{\Omega}}} \B{v}_{\text{ref}} \\
		      & -\skewsym{\nnx{\B{\Omega}}}^2 \B{p}_{\text{ref}} - \nnx{\widehat{\B{a}}})
	\end{aligned}
\end{equation}
Therefore, $\nbx{\B{z}} = \B{t} / \lVert\B{t}\rVert$.
To ensure the quadrotor flight safety, the orientation of $\nbx{\B{z}}$ is limited in a cone ($z\geq 0$) around the $z$ axis of frame $N$, i.e., $\arctan(\nbx{\B{z}}^z / \sqrt{(\nbx{\B{z}}^x)^2 + \nbx{\B{z}}^y)^2}) \in [\theta_{\text{low}}, \pi - \theta_{\text{low}}]$, where $\theta_{\text{low}} \in (0,\frac{\pi}{2})$.
Assuming the quadrotor's $x$-axis is aligned with the $x$-axis of the non-inertial frame $N$, i.e., ${}^N\B{x}_C = [1,0,0]^\top$.
We can determine $\nbx{\B{x}}$ and $\nbx{\B{y}}$:
\begin{equation}
	\label{eq:x_and_y}
	\nbx{\B{y}} = \frac{\nbx{\B{z}}\times{}^N\B{x}_C}{\lVert \nbx{\B{z}}\times{}^N\B{x}_C \rVert}, \nbx{\B{x}} = \nbx{\B{y}} \times \nbx{\B{z}}
\end{equation}
Since $\nbx{\B{z}}$ is confined within a cone, $\nbx{\B{z}}\times{}^N\B{x}_C \neq 0$ is guaranteed, and we can define the reference rotation matrix $\B{R}_{\text{ref}}$:
\begin{equation}
	\label{eq:rotation_ref}
	\B{R}_{\text{ref}} = [\nbx{\B{x}},\nbx{\B{y}},\nbx{\B{z}}]
\end{equation}
and the corresponding reference quaternion $\B{q}_{\text{ref}}$.
Based on the above arguments and assumptions, we can generate the local collision-free trajectory (Fig.~\ref{fig:concept}) that obeys the quadrotor dynamics $\nonds$, as described in Algo.~\ref{algo:genTrajectory}.
By applying the property of differential flatness, one can also generate the reference angular velocity for the trajectories.
However, for safety considerations, we choose a conservative hovering reference input $\B{u}_{\text{ref}} = \left[ g, 0, 0, 0 \right]^\top$.
In this algorithm, we assume that within each trajectory-generation window $t \in \{\Delta t, 2\Delta t,\dots,n\Delta t\}$, the non-inertial quantities $\B{\mathcal{N}}$ and the sampled data points $\B{\Phi}$ are constant. And It should be noted when $\B{\mathcal{N}} = [(0,0,g)^\top;\B{0};\B{0}]$, the quadrotor dynamics will degenerate to its state in the world frame, so this trajectory generation method naturally applies to individual quadrotors in the traditional world frame.

\begin{algorithm}[t]
	\caption{GenTtrajectory}
	\label{algo:genTrajectory}
	\SetKwInput{Input}{Input}
	\SetKwInput{Output}{Output}
	\SetAlgoLined

	\Input{Relative position estimation $\nbx{\B{p}}$, Sampled data points $\B{\Phi}$, Non-inertial quantities $\B{\mathcal{N}}$, Global goal $\B{p}_{\text{goal}}$, Trajectory size $h$, Discretized time $\Delta t$}

	\Output{Reference trajectory $\B{\Lambda}$}

	$\B{p}_{\text{last}} \leftarrow \nbx{\B{p}}$ \\
	$\B{v}_{\text{init}} \leftarrow bound(-k_p(\B{p}_{\text{last}} - \B{p}_{\text{goal}}))$ \\
	$\B{v}_{\text{last}} \leftarrow \B{\mathcal{M}}(\B{p}_{\text{last}}, \B{v}_\text{init})\B{v}_\text{init}$\\

	\For{$i \leftarrow 1$ to $h$}
	{
	$\B{p}_{\text{ref}} \leftarrow \B{p}_{\text{last}} + \B{v}_{\text{last}} \Delta t$

	$\B{v}_{\text{init}} \leftarrow  bound(-k_p(\B{p}_{\text{ref}} - \B{p}_{\text{goal}}))$

	\tcp{Get the sampled-based modulation matrix from Eq.~\ref{eq:weighted_reference_direction}-\ref{eq:weights}}

	$\B{r}(\B{p}_{\text{ref}}) \leftarrow \text{GetRefDirection}(\B{p}_{\text{ref}}, \B{\Phi})$

	$\B{E}(\B{p}_{\text{ref}}) \leftarrow \text{GenBasisMatrix}(\B{r}(\B{p}_{\text{ref}}))$

	$\B{D}(\B{p}_{\text{ref}}, \B{v}_\text{init}) \leftarrow \text{GetDiagonalMatrix}(\B{r}(\B{p}_{\text{ref}}), \B{v}_\text{init})$

	$\B{\mathcal{M}}(\B{p}_{\text{ref}}, \B{v}_\text{init}) \leftarrow \B{E}(\B{p}_{\text{ref}})\B{D}(\B{p}_{\text{ref}}, \B{v}_\text{init})\B{E}(\B{p}_{\text{ref}})^{-1}$

	\tcp{Get reference velocity and quaternion from Eq.~\ref{eq:v_init}-\ref{eq:rotation_ref}}

	$\B{v}_{\text{ref}} \leftarrow \B{\mathcal{M}}(\B{p}_{\text{ref}}, \B{v}_\text{init})\B{v}_\text{init}$

	$\B{q}_{\text{ref}} \leftarrow \text{GetRefQuaternion}(\B{p}_{\text{ref}}, \B{v}_{\text{ref}}, \B{v}_{\text{last}}, \B{\mathcal{N}}, \B{\Delta t})$

	insert $\{\B{p}_{\text{ref}}, \B{v}_{\text{ref}}, \B{q}_{\text{ref}}\}$ to $\B{\Lambda}$

	$\B{p}_{\text{last}} \leftarrow \B{p}_{\text{ref}}$

	$\B{v}_{\text{last}} \leftarrow \B{v}_{\text{ref}}$
	}
\end{algorithm}

According to Eq.~\ref{eq:weighted_reference_direction} and Algo.~\ref{algo:genTrajectory}, the computational complexity of the trajectory generation process is $\mathcal{O}(nh)$ where $n = |\B{\Phi}|$ is the number of the LiDAR points and $h = |\B{\Lambda}|$ is the trajectory length.

\subsection{Implementation \label{subsec: Implementation}}

\begin{figure}[!t]
	\setlength{\belowcaptionskip}{-0.25cm}
	\centering
	\includegraphics[width=.476\textwidth]{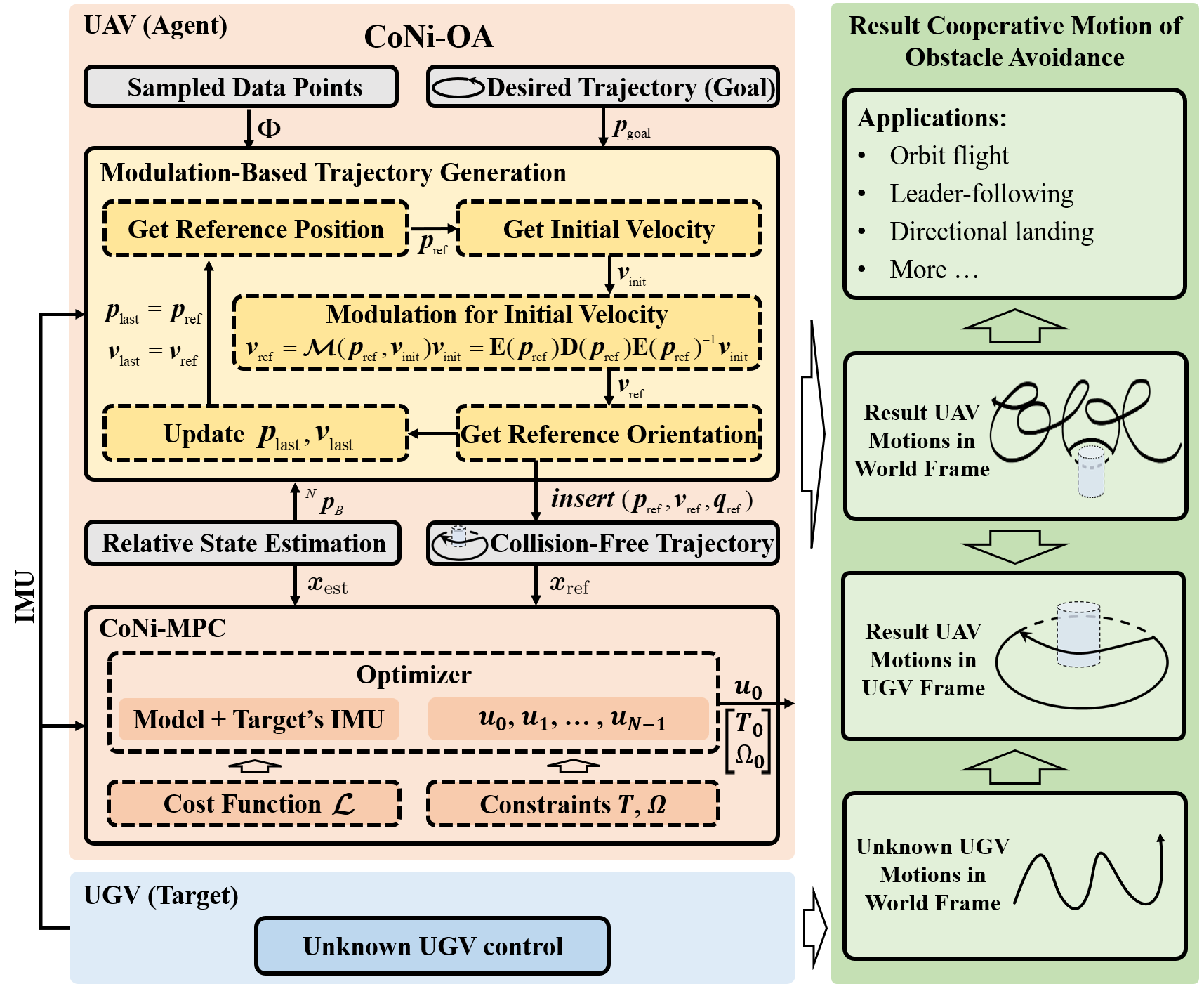}
	\caption{
		System overview and implementation of UAV-UGV cooperative obstacle avoidance using CoNi-OA.
	}
	\label{fig:system_overview}
\end{figure}

Fig.~\ref{fig:system_overview} shows the system overview of CoNi-OA for the UAV-UGV cooperative system.
Collision-free trajectories are generated by Algo.~\ref{algo:genTrajectory}, which extracts global goals from a desired relative motion trajectory referring to the UGV, using LiDAR data points and relative position estimations as inputs.
Subsequently, a CoNi-MPC controller designed based on the dynamics $\nonds$ is used to effectively track these trajectories.
The cost function of this MPC optimization problem can be defined as:
\begin{equation}
	\label{eq:cost_function}
	\begin{aligned}
		\mathcal{L} = & \sum_{k=0}^{N-1} (
		\lVert \B{x}(k) - \B{x}(k)_{\text{ref}} \rVert_{\B{Q}} +
		\lVert \B{u}(k) - \B{u}_h \rVert_{\B{R}} )                                              \\
		              & + \lVert \B{x}(N) - \B{x}(N)_{\text{ref}} \rVert_{\B{Q}_{\text{final}}}
	\end{aligned}
\end{equation}
where $\lVert \B{x}\rVert_{\B{M}} = \B{x}^\top \B{M} \B{x}$, $\B{x}(k)_{\text{ref}} = [\B{p}(k)_{\text{ref}};\B{v}(k)_{\text{ref}};\B{q}(k)_{\text{ref}}]$ is the reference state of the reference trajectory $\B{\Lambda}$ generated from Algo.~\ref{algo:genTrajectory} and $\B{u}_h = [g;\B{0}]$ is the reference input. The input $\B{u}$ is limited by the constraints:
$
	\label{eq:constraint_T}
	T_{\text{min}} \leq T \leq T_{\text{max}}
$,
$
	\label{eq:constraint_rp}
	-\Omega_{\text{rp}} \leq {}^B\Omega_B^{x,y} \leq \Omega_{\text{rp}}
$,
$
	\label{eq:constraint_yaw}
	-\Omega_{\text{yaw}} \leq {}^B\Omega_B^z \leq \Omega_{\text{yaw}}
$
where $T_{\text{min}}, T_{\text{max}},\Omega_{\text{rp}}, \Omega_{\text{yaw}} \in \mathbb{R}_{>0}$.

CoNi-MPC is implemented using the ACADO Toolkit \cite{houska2011acado}, which is employed to calculate thrust and body rate command for the UAV.
The UGV's IMU measurements, transmitted through ROS's multi-machine communication mechanism, are used to estimate $\nnx{\widehat{\B{a}}}$ and $\nnx{\B{\Omega}}$ of the non-inertial quantities $\B{\mathcal{N}}$.
For the angular acceleration $\nnx{\B{\beta}}$, we set this term to $\B{0}$ because it is hard to retrieve, thereby assuming the UGV rotates with a constant angular velocity in each prediction window. For each control iteration of CoNi-MPC, the time window is set as $\B{\textit{T}} = 2 \text{ s}$ and the discretized time is set as $\B{\textit{dt}} = 0.1\text{ s}$.
We run the real control loop at a frequency of at least 100Hz, which is much faster than the discretized time. Due to the small scale of optimization problem, the frequent update of control command by CoNi-MPC does not require much computational cost and is conducive to facilitating the convergence.

\section{Experiments}

\begin{figure}[!t]
	\setlength{\belowcaptionskip}{-0.25cm}
	\centering
	\includegraphics[width=.476\textwidth]{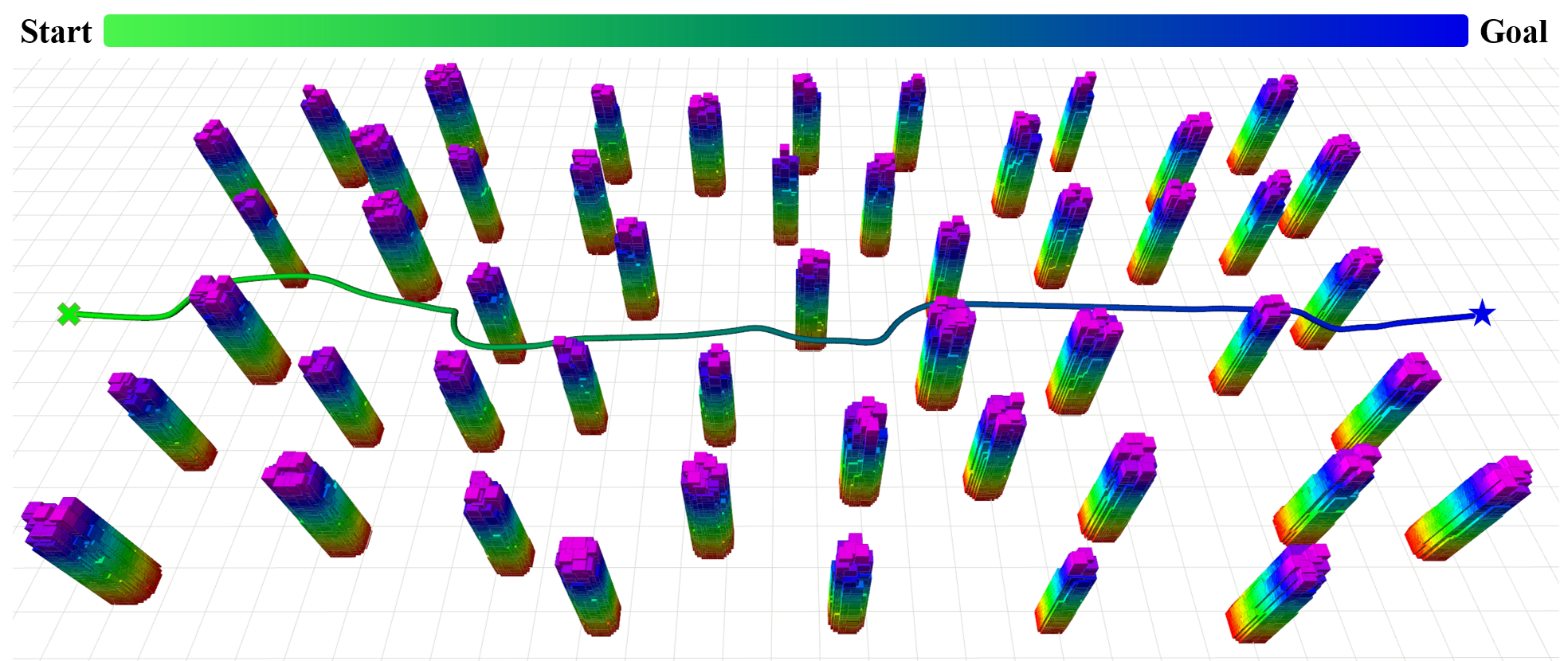}
	\caption{The collision-free trajectory of the quadrotor generated by Algo.~\ref{algo:genTrajectory} transitions from green at the starting point to blue at the goal.}
	\label{fig:collision-free}
\end{figure}

\begin{figure}[!t]
	\setlength{\belowcaptionskip}{-0.50cm}
	\centering
	\includegraphics[width=.476\textwidth]{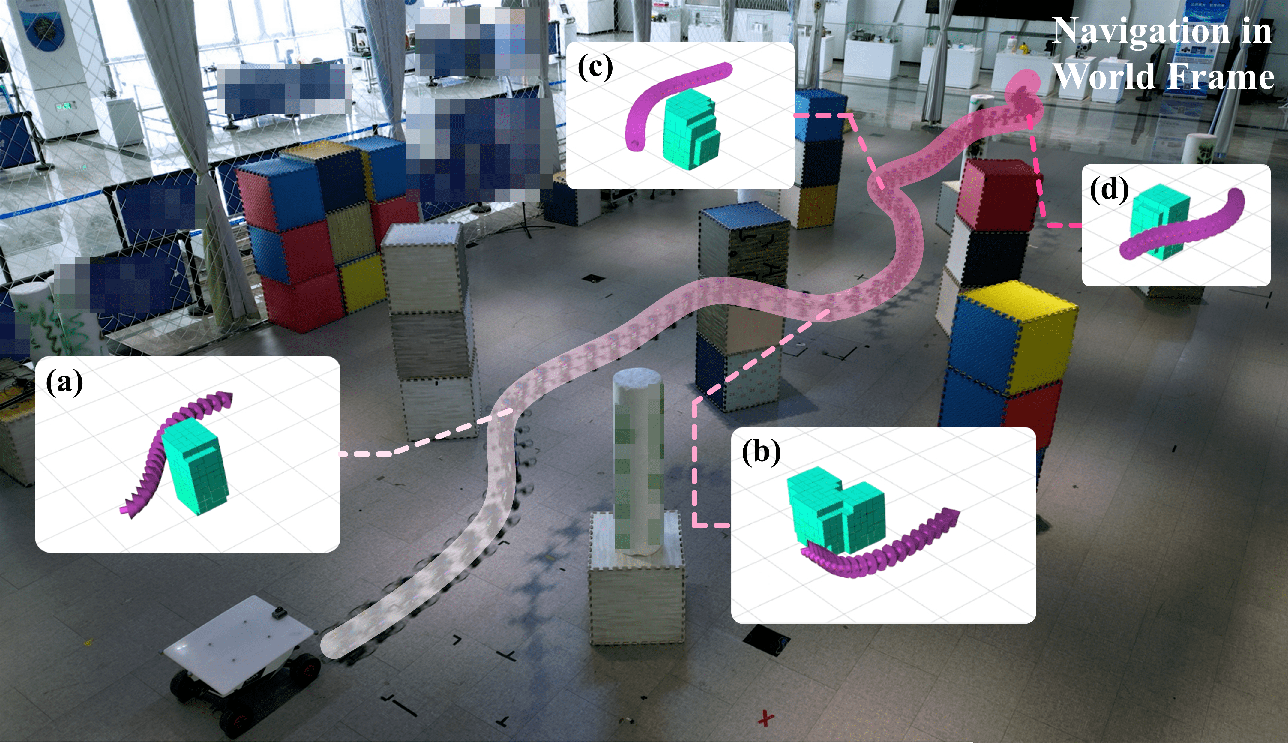}
	\caption{Obstacle avoidance flight of the quadrotor in a degenerated scenario, where the UGV keeps static as an inertial world frame. (a)-(d) show the generated collision-free trajectories when the quadrotor encounters obstacles.}
	\label{fig:conventional_avoid}
\end{figure}

To validate the effectiveness and reliability of CoNi-OA in dynamic obstacle scenarios and UAV-UGV cooperative frameworks, we conducted a comprehensive set of experiments. These include real-world demonstrations in diverse scenarios, such as leader-following, orbit flight, and directional landing, showing the practical applicability of CoNi-OA. Furthermore, we evaluated its performance against the state-of-the-art Ego-Planner \cite{zhou2020ego} in simulation to highlight its advantages. To demonstrate the versatility of CoNi-OA as a general framework for obstacle avoidance beyond dynamic environments, we performed both simulations (Fig.~\ref{fig:collision-free}) and real-world experiments (Fig.~\ref{fig:conventional_avoid}) in conventional navigation tasks, where the quadrotor operates in the world inertial frame with static obstacles.

\begin{figure}[!t]
	\setlength{\belowcaptionskip}{-0.25cm}
	\centering
	\includegraphics[width=.476\textwidth]{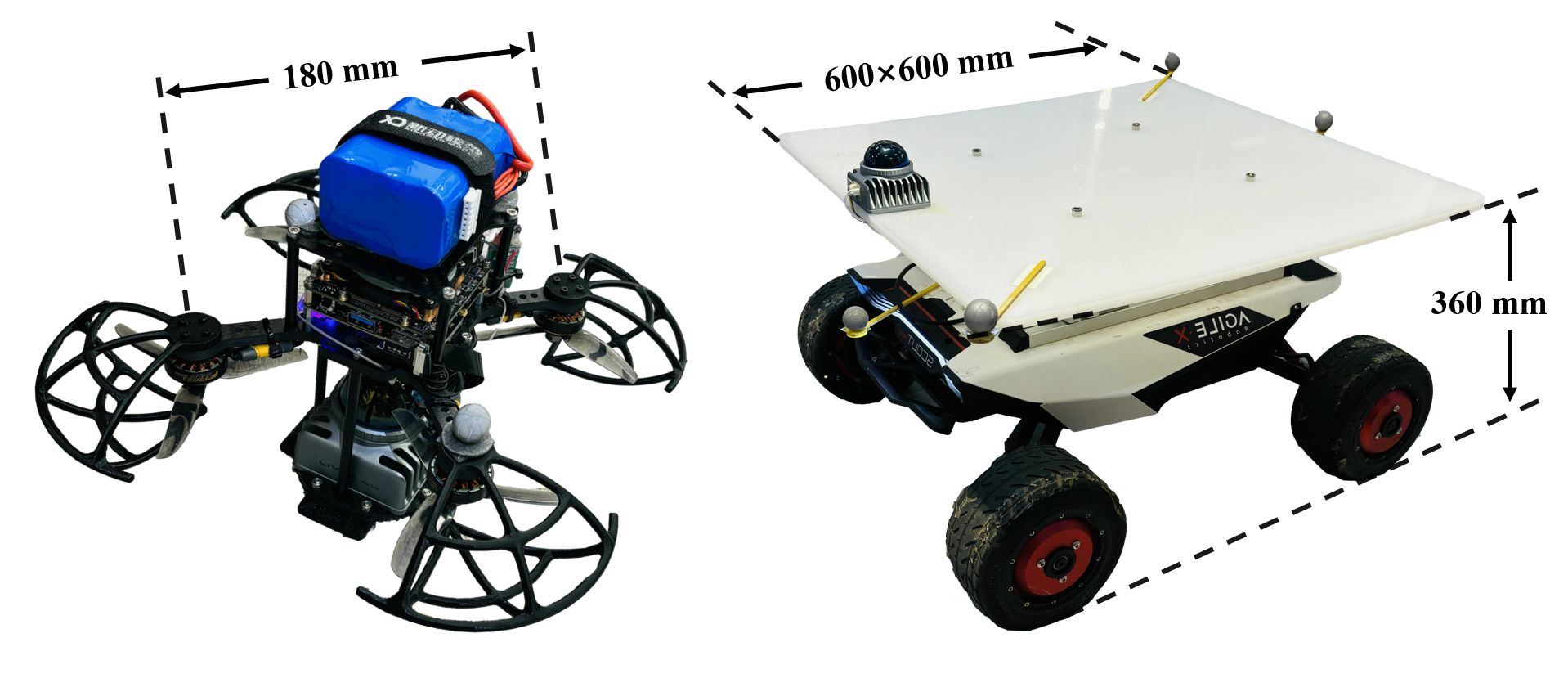}
	\caption{The quadrotor and UGV used for our experiments.}
	\label{fig:platform}
\end{figure}

\subsection{Real-World Experiments}

In order to prove the functionality of our CoNi-OA, we ran three experiments with different cooperative tasks.
In these experiments, we integrated UAV-UGV cooperative tasks, demonstrating stable obstacle avoidance during leader-following, orbit flight, and directional landing in the non-inertial frame.

The UAV and UGV platforms used are shown in Fig.~\ref{fig:platform} along with their dimensions.
The relative state estimation is computed from the NOKOV motion capture system.
The quadrotor, with a weight of 974.4g, achieves a thrust-to-weight ratio of 2.78 when powered by a 6S battery.
Trajectory generation and CoNi-MPC are implemented on an onboard computer with an Intel N100 processor (up to 3.40 GHz) and 16 GB of RAM, while the Livox Mid-360 provides the sampled data points.
The average computation time for one iteration of the controller is only 0.24 ms, which far exceeds the standard requirement for quadrotor control.
When processing a set containing about 4000 data points, it can generate a collision-free descretized trajectory of length 20 ($dt$ = 0.1 s) within an average of 4.91 ms, enabling a frequency of over 100 Hz.

\begin{figure}[!t]
	\setlength{\belowcaptionskip}{-0.50cm}
	\centering
	\includegraphics[width=.476\textwidth]{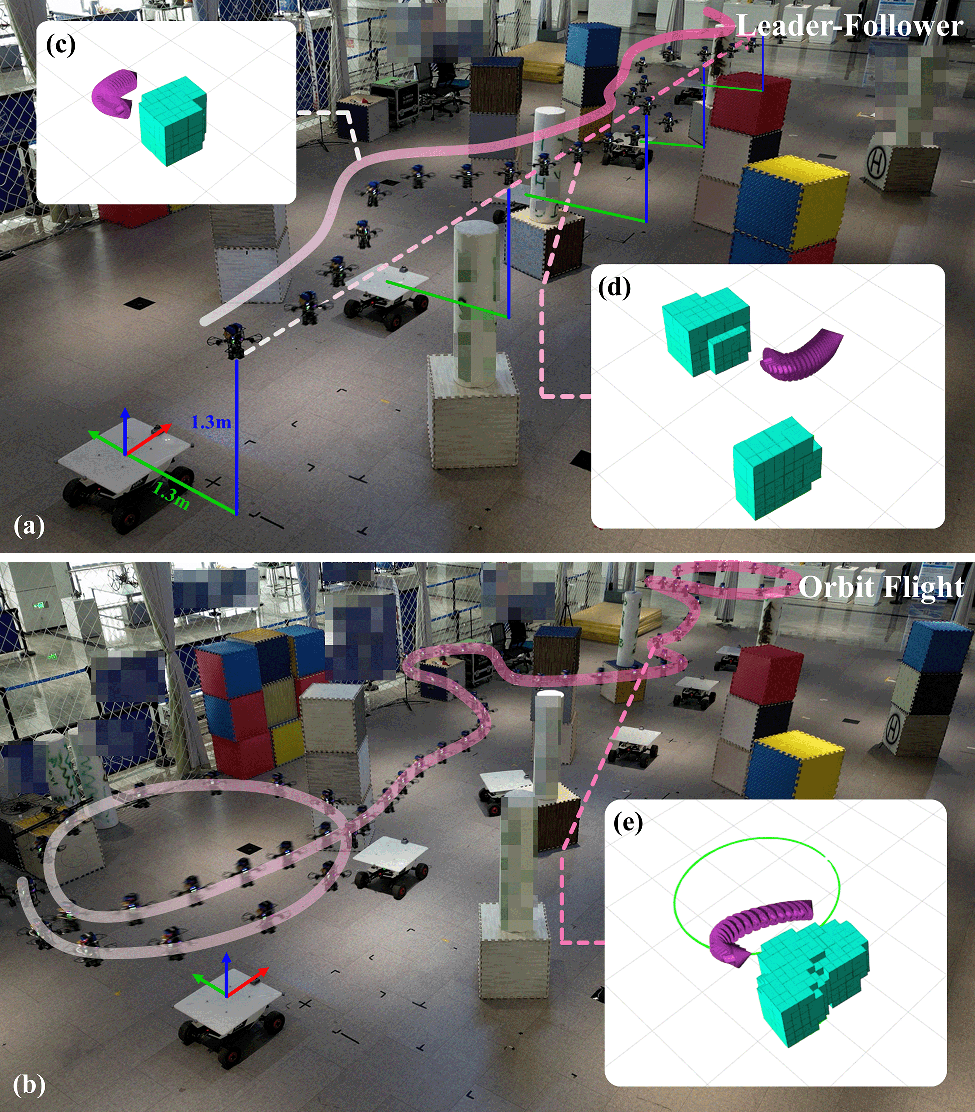}
	\caption{Obstacle avoidance flight of the quadrotor during (a) leader-following and (b) orbit-flight tasks. (c)-(e) show the generated collision-free trajectories when the quadrotor encounters obstacles.}
	\label{fig:follow_and_orbit}
\end{figure}

\begin{figure}[!t]
	\setlength{\belowcaptionskip}{-0.50cm}
	\centering
	\includegraphics[width=.476\textwidth]{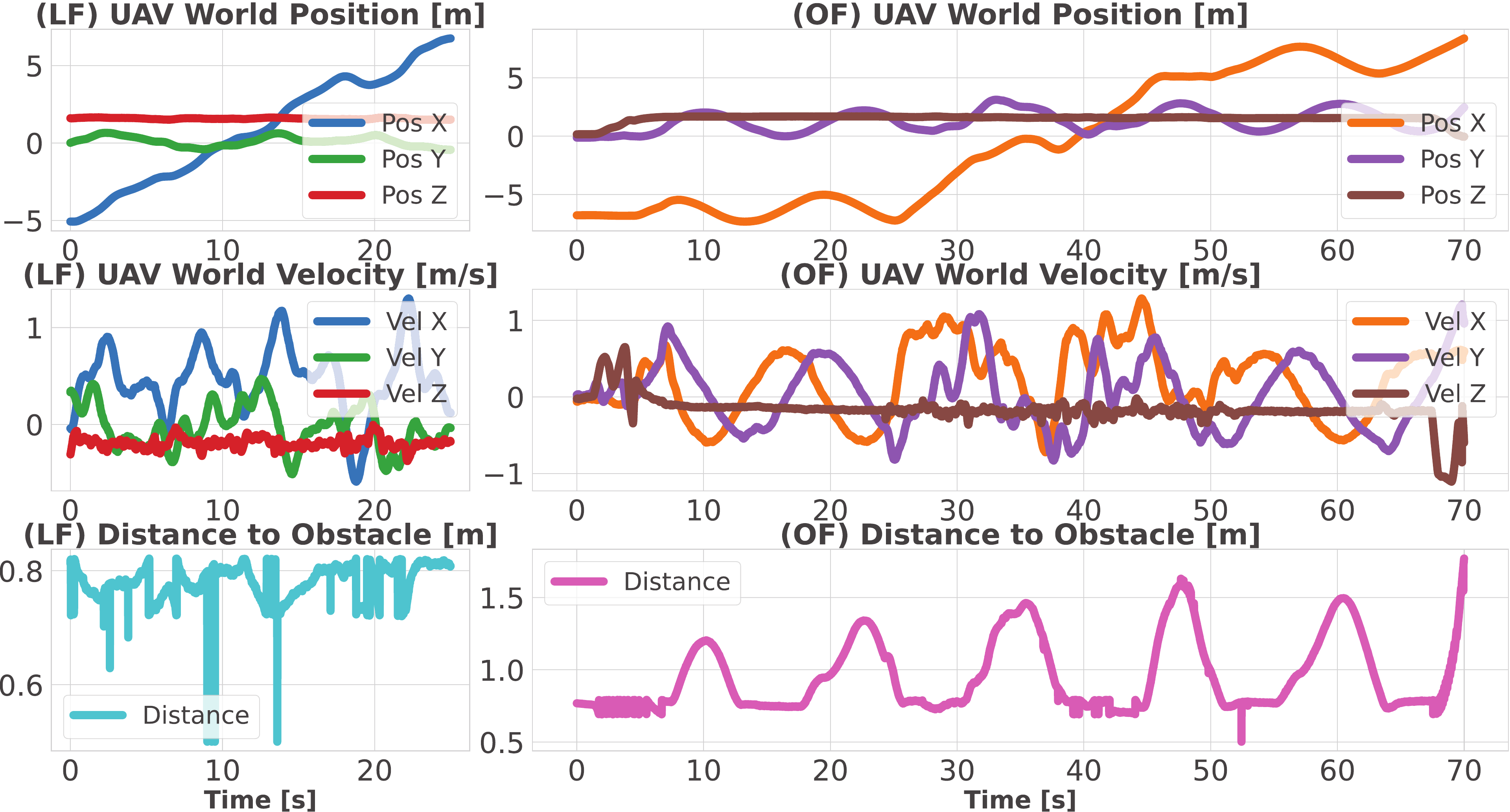}
	\caption{UAV flight data for LF and OF tasks. From top to bottom: UAV position, UAV velocity, and distance to obstacles.}
	\label{fig:world_figure}
\end{figure}

The results of cooperative obstacle avoidance in leader-following (LF) and orbit-flight (OF) tasks are shown in Fig.~\ref{fig:follow_and_orbit}.
The position, velocity, and distance-to-obstacle metrics over time for LF and OF tasks are shown in Fig.~\ref{fig:world_figure}.
The UGV utilizes Ego-Planner \cite{zhou2020ego} and an MPC controller for navigation in the world frame, while the quadrotor follows a fixed point ($\nbx{\B{p}}=(0.0, -1.3, 1.3)^{\top}$) and a circular trajectory ($r=1.5 \text{ m},\; \omega=0.5 \text{ rad/s}$) as goals, respectively. As the UGV moves, the quadrotor effectively avoids obstacles, which are dynamic relative to it, and quickly resumes the predefined trajectory.
Obstacle avoidance during the directional landing process is illustrated in Fig.~\ref{fig:landing}. The UGV rotates at an angular speed of $0.5 \text{ rad/s}$, while the quadrotor successfully lands on the platform of the UGV without colliding with any obstacles.

% \vspace{-1.0cm}

\vspace{-0.5cm}
\subsection{Comparision Study}
To evaluate the performance of CoNi-OA, we compared it against Ego-Planner \cite{zhou2020ego}, VO\cite{van2008reciprocal}, and AVO \cite{van2011avo} in simulation using ROS and Rviz.
Since Ego-Planner, VO, and AVO were originally developed for the inertial frame, in this comparison study we used the same relative observations (LiDAR data, positions, velocities, and orientations) as those in CoNi-OA and CoNi-MPC for both methods.
Specifically, for Ego-Planner we replaced its original depth sensor with a local grid map simulating LiDAR sensors.
For VO and AVO, we also utilized a local grid map; however, instead of employing dedicated obstacle detection methods, each point was treated as a small obstacle to estimate its velocity and acceleration.
All methods ran on a desktop computer equipped with an Intel i7-12600KF processor (up to 4.90\,GHz) and 32\,GB of RAM, using CoNi-MPC \cite{zhang2023coni} as the underlying controller with identical parameters.
We examined two tasks: leader-follower (LF) and orbit flight (OF). In the LF task, the UAV tracked a fixed position \((1.0, 0.0, 0.5)\) relative to the UGV; in the OF task, it followed a predefined circular trajectory of radius 1.0 and angular velocity 0.5, centered at \((1.0, 0.0, 0.5)\). Throughout the simulation, the UGV moved unpredictably with random goal assignments, effectively creating dynamic obstacles from the UAV's viewpoint.
We tested the approaches in a $26\text{m}\times20\text{m}\times3\text{m}$ map under varying obstacle densities: sparse (100 obstacles), medium (150 obstacles), and dense (200 obstacles), and under two different UGV motion constraints: slow movements \((v_{\max} = 0.5,\, \omega_{\max} = 0.5)\) and fast movements \((v_{\max} = 1.5,\, \omega_{\max} = 1.5)\). Table~\ref{tab:success_rate} provides the success rates for each method under these conditions.
The success rate was computed by treating each UGV navigation attempt as one trial (a total of 100 trials). During a trial, if the UAV came within 0.1\,m of any obstacle at any point, the trial was deemed a failure; otherwise, it was considered a success.
The results show that CoNi-OA achieved a higher success rate for both LF and OF tasks under various obstacle densities and UGV motion conditions.
\begin{table}[t]
	\caption{Success rates for CoNi-OA, Ego-Planner, VO, and AVO on LF and OF tasks with sparse, medium,
		and dense obstacles, under two UGV motion constraints. Each entry is the percentage
		of successful trials (out of 100), where success requires the UAV to remain more than 0.1\,m
		away from obstacles.}
	\label{tab:success_rate}
	\setlength{\belowcaptionskip}{-1.00cm}
	\centering
	\begin{adjustbox}{width=.476\textwidth}
		\begin{tabular}{c c c c c c c c c c}
			\toprule
			\multicolumn{2}{c}{\multirow{4}{*}{{Scenario}}} & \multicolumn{8}{c}{Success Rate $[\%]$}                                                                                                                                                                                                                                              \\
			\cmidrule(lr){3-10}
			                                                &                                         & \multicolumn{4}{c}{Slow UGV Movements} & \multicolumn{4}{c}{Fast UGV Movements}                                                                                                                                                            \\
			\cmidrule(lr){3-6} \cmidrule(lr){7-10}
			                                                &                                         & \textbf{CoNi-OA}                       & Ego-Planner\cite{zhou2020ego}          & VO\cite{van2008reciprocal} & AVO\cite{van2011avo} & \textbf{CoNi-OA} & Ego-Planner\cite{zhou2020ego} & VO\cite{van2008reciprocal} & AVO\cite{van2011avo} \\
			\toprule
			\multirow{3}{*}{\makecell[c]{LF}}               & Sparse                                  & \textbf{94.0}                          & 66.0                                   & 62.0                       & 61.0                 & \textbf{78.0}    & 48.0                          & 51.0                       & 38.0                 \\
			                                                & Medium                                  & \textbf{82.0}                          & 58.0                                   & 19.0                       & 32.0                 & \textbf{57.0}    & 28.0                          & 38.0                       & 23.0                 \\
			                                                & Dense                                   & \textbf{58.0}                          & 39.0                                   & 9.0                        & 28.0                 & \textbf{24.0}    & 14.0                          & 20.0                       & 12.0                 \\
			\hline
			\multirow{3}{*}{\makecell[c]{OF}}               & Sparse                                  & \textbf{72.0}                          & 31.0                                   & 56.0                       & 49.0                 & \textbf{53.0}    & 25.0                          & 45.0                       & 47.0                 \\
			                                                & Medium                                  & \textbf{58.0}                          & 13.0                                   & 25.0                       & 42.0                 & \textbf{32.0}    & 11.0                          & 24.0                       & 14.0                 \\
			                                                & Dense                                   & \textbf{36.0}                          & 11.0                                   & 6.0                        & 18.0                 & \textbf{21.0}    & 8.0                           & 15.0                       & 9.0                  \\
			\toprule
		\end{tabular}
	\end{adjustbox}
\end{table}

\section{Conclusion}

In this paper, we present a modulation-based, non-inertial-frame cooperative obstacle avoidance algorithm, CoNi-OA, designed to address safety challenges in air-ground collaborative systems. The UAV is directly controlled in the UGV's non-inertial frame, eliminating reliance on global states, thus enabling effective cooperative obstacle avoidance during specific UAV-UGV interaction tasks. The proposed approach iteratively generates collision-free trajectories using only a single frame of LiDAR data points without explicit obstacle detection or prediction. These trajectories comply with non-inertial quadrotor dynamics and are efficiently tracked by CoNi-MPC. CoNi-OA provides a general solution for obstacle avoidance applicable to both static and dynamic environments, and while primarily intended for quadrotors in air-ground collaborative systems, it also offers a robust navigation solution in global frames.
Through extensive simulation studies and real-world experiments, the effectiveness and robustness of CoNi-OA have been demonstrated. Nevertheless, we identified certain limitations of the proposed approach. Due to the relatively low update frequency of LiDAR data points, the method can struggle to generate timely collision-free trajectories, particularly during aggressive maneuvers of the UGV.
In the future, we aim to improve CoNi-OA by integrating high-frequency LiDAR sensors, extend it to multi-robot cooperative obstacle avoidance scenarios, and incorporate real-time relative estimation devices to enhance its practicality and reliability.

\bibliographystyle{IEEEtran}
\bibliography{paper}

\end{document}